\documentclass[letterpaper]{article} % DO NOT CHANGE THIS
\usepackage[submission]{aaai23}  % DO NOT CHANGE THIS
\usepackage{times}  % DO NOT CHANGE THIS
\usepackage{helvet}  % DO NOT CHANGE THIS
\usepackage{courier}  % DO NOT CHANGE THIS
\usepackage[hyphens]{url}  % DO NOT CHANGE THIS
\usepackage{graphicx} % DO NOT CHANGE THIS
\usepackage{natbib}  % DO NOT CHANGE THIS AND DO NOT ADD ANY OPTIONS TO IT
\usepackage{caption} % DO NOT CHANGE THIS AND DO NOT ADD ANY OPTIONS TO IT
\usepackage{algorithm}
\usepackage{algorithmic}
\usepackage{amsmath} % for equation aligned
\usepackage{url}
\usepackage{newfloat}
\usepackage{listings}
\DeclareCaptionStyle{ruled}{labelfont=normalfont,labelsep=colon,strut=off} % DO NOT CHANGE THIS
\floatstyle{ruled}
\newfloat{listing}{tb}{lst}{}
\floatname{listing}{Listing}
\usepackage{booktabs}  
\usepackage{multirow}
\usepackage{threeparttable}
\usepackage{bm}
\usepackage{float}
\usepackage{amsmath,amsfonts}
\usepackage{makecell}
\usepackage{pifont}    % 勾和叉
\usepackage{color}     % 颜色
\title{USR: Unsupervised Separated 3D Garment and Human Reconstruction via Geometry and Semantic Consistency}
\author{
	Yue Shi, \textsuperscript{\rm 1}
	Yuxuan Xiong, \textsuperscript{\rm 1}
	Jingyi Chai, \textsuperscript{\rm 1}
	Bingbing , \textsuperscript{\rm 1}
	Wenjun Zhang \textsuperscript{\rm 1}
}
\affiliations{
	\textsuperscript{\rm 1} Shanghai Jiaotong University \\
	shiyue001@sjtu.edu.cn,
	xiongyx@sjtu.edu.cn,
	chaijingyi@sjtu.edu.cn,
	nibingbing@sjtu.edu.cn,
	zhangwenjun@sjtu.edu.cn
}

\usepackage{bibentry}
\begin{document}

\maketitle

\begin{abstract}
Dressed people reconstruction from images is a popular task with promising applications in creative media and game industry. However, most existing methods reconstruct the human body and garments as a whole with the supervision of 3D models, which hinders the downstream interaction tasks and requires hard-to-obtain data. To address these issues, we propose an unsupervised separated 3D garments and human reconstruction model (USR), which reconstructs the human body and authentic textured clothes in layers without 3D models. More specifically, our method propose a generalized surface-aware neural radiance field to learn the mapping between sparse multi-view images and geometries of the dressed people. Based on the full geometry, we introduce a Semantic and Confidence Guided Separation strategy (SCGS) to detect, segment, and reconstruct the clothes layer, leveraging the consistency between 2D semantic and 3D geometry. Moreover, we propose a Geometry Fine-tune Module to smooth edges. Extensive experiments on our dataset show that comparing with state-of-the-art methods, USR achieves improvements on both geometry and appearance reconstruction while supporting generalizing to unseen people in real-time. Besides, we also introduce SMPL-D model to show the benefit of the separated modeling of clothes and human body that allows swapping clothes and virtual try-on.
\end{abstract}

\section{Introduction}
With the promotion of digital human models in virtual try-on, game industry, and creative media, the separated reconstruction of human body and clothing becomes urgently needed for its advantage on trying-on and interaction of human and garment in the real world. However, existing separated reconstruction methods rely on the supervision of 3D ground truth and limited clothing templates, which need manually modeled and annotated 3D mesh and cannot cope with the topological complexity of various garments. Therefore, it is meaningful to propose an unsupervised separated reconstruction framework (USR) to reconstruct 3D garment and human body from readily available 2D images. 

%备用的表格1

\begin{table}[t]
\centering
\resizebox{.48\textwidth}{!}{
\setlength\tabcolsep{1.2pt}{
\renewcommand\arraystretch{1.5}
\begin{tabular}{m{3.5cm}<{\centering}|m{1.27cm}<{\centering}m{1.82cm}<{\centering}m{1.82cm}<{\centering}m{1.6cm}<{\centering}}
%m{2cm}<{\centering}|m{1cm}<{\centering} m{1cm}<{\centering} m{1cm}<{\centering}m{1cm}<{\centering}m{1cm}<{\centering}
%\toprule
\hline\hline
Methods & Without 3D Sup. & Separating Human clothes & Arbitrary clothes Topology & Realistic appearance \\
%\midrule
\hline
PIFu \cite{saito2019pifu}, PIFuHD \cite{saito2020pifuhd}, SCANimate \cite{saito2021scanimate} & \textcolor{red}{\ding{55}} & \textcolor{red}{\ding{55}} & \textcolor{red}{\ding{55}} & \textcolor{red}{\ding{55}}  \\
SiCloPe \cite{natsume2019siclope} & \textcolor{green}{\checkmark} & \textcolor{red}{\ding{55}} & \textcolor{red}{\ding{55}} & \textcolor{green}{\checkmark} \\
MGN \cite{mgn}, Octopus \cite{Octopus_2019_CVPR}, Tex2Shape \cite{tex2shape} & \textcolor{red}{\ding{55}} & \textcolor{green}{\checkmark} & \textcolor{red}{\ding{55}} & \textcolor{green}{\checkmark} \\
SMPLicit \cite{corona2021smplicit} & \textcolor{red}{\ding{55}} & \textcolor{green}{\checkmark} & \textcolor{green}{\checkmark} & \textcolor{red}{\ding{55}} \\
Ours  & \textcolor{green}{\checkmark} & \textcolor{green}{\checkmark} & \textcolor{green}{\checkmark} & \textcolor{green}{\checkmark}\\
%\bottomrule
\hline\hline
\end{tabular}
}
}
\vspace{-0.2cm}
\caption{Characteristic comparison of our model against previous clothing human reconstruction methods from 2D monocular images.}
\label{different}
\vspace{-0.5cm}
\end{table}

Traditional clothing human reconstruction algorithms can be divided into two categories according to the input data format, i.e., scan-based and image-based. Scan-based reconstruction methods have achieved significant results \cite{saito2021scanimate,saito2020pifuhd}, achieving accurate 3D models and even precise fold details. However, scan-based methods require expensive and bulky equipment, which are not friendly to general users. Image-based methods are represented by PIFu \cite{saito2019pifu}, PIFuHD \cite{saito2020pifuhd}, and Tex2Shape \cite{tex2shape}, using one image to recover the 3D model leveraging the learned commonalities between humans. Octopus \cite{Octopus_2019_CVPR} learns from multiple images, predicting the parameters of instance displacements that add clothing to the body shape. But the learning of the human prior requires enough pairs of 3D models and 2D images which are built artificially. To alleviate the lack of paired data, unsupervised human reconstruction methods \cite{doubelfield,avatarnerf} are proposed with the introduction of Neural Radiance Field (NeRF) \cite{mildenhall2020nerf}. However, all of the above methods reconstruct 3D human model as a whole, ignoring the fact that the human body is separated from the clothing in actual situations, which hinders downstream application tasks, such as changing clothes and human-clothing interaction. \textbf{Hence, separated reconstruction models for dressed people from images are urgently needed.}

% To reconstruct garment separately, Inverse Simulation \cite{guo2021inverse} represents the cloth geometry using a dynamical system that is controlled by the body states, trained by sequences of dense point clouds. Similar work is TailorNet \cite{tailornet_2020_CVPR}, which predicts clothing deformation in 3D as a function of pose, shape and garment geometry, with inputting 3D models. Both the above two methods depend on inputting relatively complete 3D models, which are difficult to obtain. 
To reconstruct from 2D images, MGN \cite{mgn} learns per-garment models from 3D scans of people in clothing and reconstructs them from images during inference stage, leveraging garment templates which are obtained from offsets of the SMPL bodies. SMPLicit \cite{corona2021smplicit} proposes a generative model for clothed bodies that can be controlled by a low-dimensional and interpretable vector of parameters. However, SMPLicit depends on 3D ground truth and is unable to handle geometric details and textures. Despite the recent work \cite{xiang2021modeling} has obtained high-fidelity two-layer mesh, it requires tens of thousands of frames strictly captured by an expensive multi-view capture system consisting of around 140 cameras which are distributed uniformly. \textbf{Therefore, challenges still remain in recovering both geometry and appearance of full people, garments, and human body from only several images without the supervision of 3D models and any labels.}

To address the issues above, we propose an unsupervised separated reconstruction framework (USR), aggregating both 2D pixel-level and semantic-level information to realize 3D layered reconstruction without 3D ground truth. Leveraging the law of multi-view geometric projection, USR learns common geometry prior from multiple people while mastering the mapping between 2D images and corresponding 3D distribution. The proposed framework is shown in Fig.~\ref{fig1}, which contains a surface reconstruction module, a separating module, and a fine-tune module. First, we propose a generalized surface-aware neural radiance field to learn the 3D geometry and appearance from multi-view image, which makes NeRF break through the limitation of single scene optimization, massive input images, and uneven surface. Then, based on the reconstructed full people, we derive a semantic label for every vertex of the geometry by fusing corresponding 2D semantic labels on multi-view images according to their confidence. Finally, we introduce a fine-tune module to smooth the edge. Extensive experiments on our collected Open-source Dressed People dataset prove that our unsupervised method even outperforms existing state-of-the-art supervised models. We also introduce SMPL-D model to show the try-on application using our reconstructed garments models and human bodies, demonstrating the great benefit of separated modeling of 3D garments and human. The main contributions of this work can be summarized as: 

1) We propose the first unsupervised separated 3D garment and human body reconstruction framework (USR); 

2) We design a generalized unsupervised 3D surface reconstruction network, extending the neural radiance field to a generalized and surface-aware manner with sparse inputs; 

3) We introduce Semantic and Confidence Guided Strategy (SCGS) for 3D garment detection, segmentation, and reconstruction without 3D segmentation labels; 

4) We propose a Geometry Fine-tune Module containing an edge smoothing algorithm and a strategy preventing penetration between layers, which can also be applied in other mesh processing tasks.

\section{Related Work}
\subsubsection{3D Human Reconstruction}
3D human reconstruction is a popular issue in computer vision. Some methods \cite{pons2017clothcap,saito2021scanimate,guo2021inverse,corona2021smplicit,ma2021scale,ma2021power,hong2021garment4d} reconstruct the 3D models of humans in clothes from 3D point clouds, which is more difficult to collect, thus hindering their wide applications. By contrast, image-based methods \cite{natsume2019siclope, bhatnagar2019multi, Octopus_2019_CVPR, saito2019pifu, saito2020pifuhd, hong2021stereopifu, alldieck2019tex2shape, jafarian2021learning, xiang2021modeling} are not limited by collection equipment and have also achieved significant progress. However, to obtain a high-resolution result, the training often requires 3D models as ground truth, which are expensive for the complex and time-consuming producing process. Thus, several attempts \cite{natsume2019siclope, jafarian2021learning, xiang2021modeling} have been made in unsupervised or weakly-supervised 3D human reconstruction. Yet \cite{natsume2019siclope, jafarian2021learning} perform regretfully to some extent while \cite{xiang2021modeling} has to depend on dense multi-view videos captured by an expensive system consisting of around 140 cameras.

% Unisurf[44] then fix this problem by reducing the sampling area gradually as well as applying constraints in normal into the model to create more subtle reconstructed surfaces.Yet the original NeRF can only be optimized on one scene at a time, which cannot be generalized to unseen scenes and is inefficient. Some works \cite{pixelnerf,mvsnerf} make the NeRF generalizable by learning the commonalities between scenes from image features. 
% Though achieving   So [29-39] focus on enhancing the capability of generalization and achieve in predicting multi or novel scenes with one trained model. Apart from that, the conventional NeRF method fails on dynamic occasion. [40] tackle this issue by additionally introducing a deformation field, besides the canonical field, to describe points' movements in a certain period. Furthermore, 

% [17] first came up with NeRF, a promising method of view synthesis by implicitly representing a 3D scene through a fully-connected deep network. Many research are then conducted based on that. [18-28] increase the efficiency by speeding up either the rendering or the training process. 

% \vspace{-0.2cm}
\subsubsection{Neural Radiance Field}
Neural Radiance Field (NeRF) \cite{mildenhall2020nerf} brings new inspiration to unsupervised 3D reconstruction, inspiring a series of novel works on unsupervised 3D human reconstruction \cite{doubelfield,avatarnerf}. NERF uses hundreds of pictures to parameterize the 3D distribution of the scene with the MLP network, which cannot deal with the problem of reducing the number of training pictures and generalizing to new scenes. To make it generalizable, some works \cite{yu2021pixelnerf,chen2021mvsnerf} make the NeRF generalizable by learning the commonalities between scenes from image features. However, these methods mainly learn to interpolate novel view images with inputting adjacent images, being poor in recovering geometric shapes due to the absence of surface constraints. In our method, we will extend NeRF to a generalized and surface-aware manner with sparse input images, learning the geometry prior from various 3D dressed people and mapping the 2D semantic and appearance to 3D models.

\subsubsection{3D Garments Reconstruction}
As digital humans are increasingly used in games and virtual try-on, it is no longer enough to reconstruct the human body as a whole. Reconstructing 3D garments and human body separately is urgently needed. To reconstruct garments separately, Inverse Simulation \cite{guo2021inverse} represents the cloth geometry using a dynamical system that is controlled by the body states, trained by sequences of dense point clouds. Similar work is TailorNet \cite{tailornet_2020_CVPR}, which predicts clothing deformation in 3D as a function of pose, shape, and garment geometry, with inputting 3D models. Both the above two methods depend on inputting relatively complete 3D models, which are difficult to obtain. To separately reconstruct garments from dressed people images, most methods \cite{corona2021smplicit, lahner2018deepwrinkles,jin2020pixel} manage to deform the pre-defined clothes templates under the supervision of 3D models, which limits the topology of various garments. We exploit the geometric correspondence between 2D and 3D, combining semantic prior of the human body, to introduce a Semantic and Confidence Guided Separation strategy (SCGS) for unsupervised 3D garment reconstruction.
% Joint semantic-prior and confidence segmentation strategy (JSCS)

\section{Methodology}
\begin{figure*}[h]
\centering
\includegraphics[width=18cm]{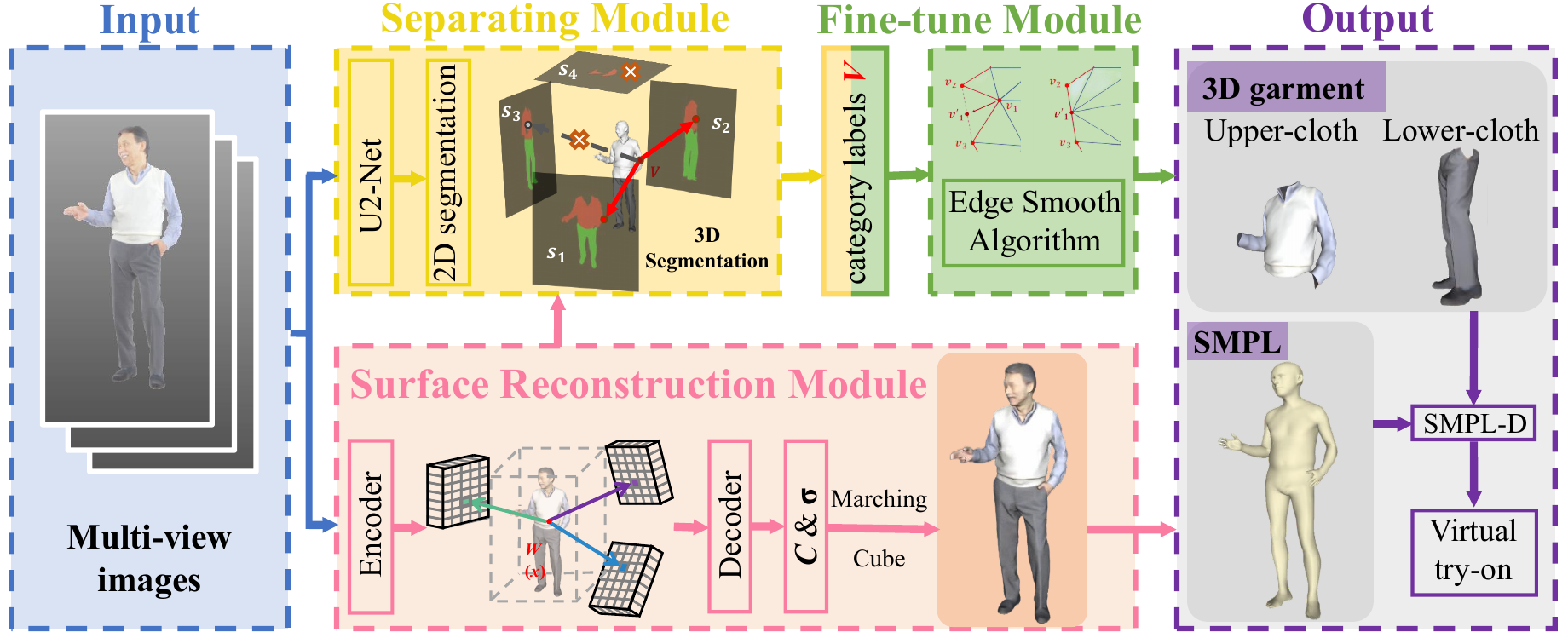}
\setlength{\belowcaptionskip}{-0.5cm}
\vspace{-0.5cm}
\caption{Illustration of the geometry-aware dynamic sampling.}
\label{fig1}
\end{figure*}

%别的方法有什么问题，
Existing 3D dressed people reconstruction methods \cite{saito2019pifu,mgn} typically utilize the  distance supervision between the predicted 3D distribution and the ground truth to learn the mapping from 2D images to 3D models, requiring hard-to-obtain 3D meshes which are manually modeled. Besides, most reconstruction methods ignore the multi-layered structure of the dressed people, which hinders their flexible use in downstream tasks. To circumvent the above problems, we design an Unsupervised Separated 3D Garment and Human Reconstruction framework (USR) to reconstruct 3D garments and human bodies separately with only multi-view RGB images as input. The overall pipeline is shown in Fig.~\ref{fig1}, which consists of three novel components, i.e., a Generalized Surface-aware Neural Radiance Field (GSNeRF) for surface reconstruction, a Semantic and Confidence Guided Separation strategy (SCGC) to separately reconstruct 3D garments, and a Fine-tune Module for edges refinement. In addition, we introduce SMPL-D model to register the reconstructed garments to other human bodies reasonably, which supports the virtual try-on.
% Joint semantic-prior and confidence segmentation strategy (JSCS) for the Separation of 3D Garment, a geometry fine-tune module, and a parametric dressing model SMPL-D.

\subsection{Surface Reconstruction Module}

\subsubsection{Generalized Surface-aware NeRF}
Original NeRF \cite{zhao2017pyramid} is proposed for novel-view synthesis, leveraging  to be optimized. Though NeRF has achieved impressive results, its requirements for hundreds of dense inputting images and single-scene training is not suitable for applying in 3D dressed people reconstruction. Even some generalized NeRFs \cite{yu2021pixelnerf,wang2021ibrnet,chen2021mvsnerf} have been proposed, they mainly focus on the interpolation of adjacent viewing angles, incapable of surface reconstruction due to the lack of 3D constraints. Therefore, we propose a Generalized Surface-aware Neural Radiance Field (GSNeRF) to learn the geometry and appearance distribution prior from 3D points features, and inference the shape and color of the unseen 3D people with only several multi-view images captured randomly. GSNeRF reconstructs 3D dressed people models with high-fidelity geometry and appearance, preparing for subsequent body and garments reconstruction.     

Different from the original NeRF, GSNeRF adds the features of the 3D points to learn the commonalities from different images of various dressed people. Specifically, for a 3D point $x\in R^{3}$, we derive its feature $W(x)$ by reflecting it to feature volumes of multi-view images along the shooting direction and fusing multi-view feature vectors. Then, combining the 3D point coordinate, the feature, and target viewing direction unit vector $d\in R^{3}$, the neural network $f_{GSNeRF}$ returns a differential density $\sigma$ and RGB color $c$, which can be described as follows:
\begin{align}
\label{network}
f_{GSNeRF}:\left ( W(x),\gamma(x),d \right )\mapsto \left ( c,\sigma  \right ),
\end{align}
where $\gamma(\cdot)$ is a positional encoding on x with 6 exponentially increasing frequencies introduced in the original NeRF \cite{kosiorek2021nerf}. Let $O$ denote the location of the camera center. Given $N$ samples points ${x_i}$ along ray $r = O + td$, we predict their density and color by $f_{GSNeRF}$ and render target 2D the pixel using the volumetric rendering:
\begin{align}
  \label{render}
  \hat{C}=\sum_{j=1}^{N}T_{j}\left ( 1-exp(-\sigma _{j}\delta_{j}) \right )c_{j}, 
\end{align} 
\vspace{-0.5cm}
\begin{align}
  T_{j}= exp(-\sum_{k=1}^{j-1}\sigma _{k}\delta_{k}),
\end{align}
where $T_{j}$ is the accumulated transmittance along the ray and $\delta_{j}$ is the distance between adjacent samples $x_j$ and $x_{j+1}$. Then we decompose and refactor Eq.(\ref{render}) to Eq.(\ref{physical}), making its physical meaning clearer. The specific derivation process is provided in the supplementary material.
\begin{align}
  \label{physical}
  \hat{C}=\sum_{j=1}^{N}o_{j}(x_j)\prod_{k=1}^{j-1}(1-o_{k}(x_k))c_j,
\end{align} 
where $o(x_i)=1-exp(-\sigma_i\delta_i)$, indicating the occupancy of $x_i$. In the novel-view synthesis task, all points ${x_i}$ on the ray participate in the rendering as Eq.(\ref{render}). There are inevitably some points that are not on the geometric surface, which are beneficial to satisfy the 2D image constraints, but affect the synthesis of accurate geometry. In surface reconstruction, we expect that only the surface point contributes to the rendered color, i.e., we expect o take 0 in free space and 1 in occupied space, acting as an occupancy indicator. Then the $\hat{C}$ equals to the color of the first occupied sample and equals to 0 before this sample. By restricting the value of o, we make the radiance field aware of the surface and eliminate the contribution of non-surface sampling points to the rendered color, allowing GSNeRF to learn implicit surfaces, i.e. geometry shape.
% In practice, we found it more beneficial to directly predict the occupancy $o$ of each sample than predict the density $\sigma$. As a result, we change equation \ref{network} into equation \ref{network2}
% \begin{align}
% \label{network2}
% f_{GSNeRF}:\left ( W(x),\gamma(x),d \right )\mapsto \left ( c, o  \right ),
% \end{align}

\subsubsection{Optimization Objectives}
We use multi-view 2D images to train the network $f_{GSNeRF}$. The first constraint is the construction loss $\mathcal{L}_{r e c}$ which measures the pixel-level similarity between the rendered target-view image and its corresponding ground truth, which can be formulated as:
% 为了公式和序号能放在一行，这里采用了小一点的字体
\begin{small}
\begin{align}
\label{recLoss}
\mathcal{L}_{r e c}=\sum_{\mathbf{r} \in \mathcal{R}}\left [\left\|\hat{C}_{c}(\mathbf{r})-C(\mathbf{r})\right\|_{2}+\left\|\hat{C}_{f}(\mathbf{r})-C(\mathbf{r})\right\|_{2}\right ],
\end{align}
\end{small}
%loss
%从method标题到这里1页
where $R$ is the set of rays in each training batch. We adopt the two-stage sampling strategy of NeRF \cite{mildenhall2020nerf}. The $\hat{C}_{c}$ and $\hat{C}_{f}$ are the predicted results of the coarse sampling stage and the fine sampling stage, respectively.

However, $\mathcal{L}_{rec}$ is weakly constrained on the surface-level. To avoid some spatial points falling out of the overall implicit surface distribution due to a lack of 2D constraints, we introduce a normal loss $\mathcal{L}_{norm}$ to ensure the implicit surface smooth, which is described as follows: 
 \begin{align}
 \label{regLoss}
     \mathcal{L}_{norm}=\sum_{\mathbf{x} \in S}\left\|\mathbf{n}\left(\mathbf{x}\right)-\mathbf{n}\left(\mathbf{x}+\epsilon\right)\right\|_{2},
 \end{align}
where $S$ is the set of the surface points. $\epsilon$ is a random perturbation in 3D space, which is gradually reduced during training. The normal $\mathbf{n}\left(\mathbf{x}\right)$ of each point $x$ is calculated by the opposite direction of the gradient of the $\sigma$ at $x$. Finally, the optimization objectives of GSNeRF are the weighted addition of the reconstruction loss and the normal loss:
 \begin{align}
     \label{normal}
     \mathcal{L}=\mathcal{L}_{rec}+\lambda \mathcal{L}_{norm}.
 \end{align} 
\subsubsection{Surface Reconstruction} 
After training, the built generalized surface-aware neural radiance field is capable of inferring the 3D distribution of an unseen person through the given multi-view images. Normally, NeRF \cite{mildenhall2020nerf} infers the color and density of a series of query points, filters out a set $S$ of surface points using a threshold, and then utilizes Marching Cube \cite{marchingcube} algorithm to extract the explicit 3D mesh. The color of each extracted vertex can be calculated by volumetric rendering along a ray with opposite direction to its normal. In the transformation, the normal of each 3D point is calculated by its neighbor points in the set $S$, the change of which is relatively drastic, comparing with our defined normal in Eq.(\ref{normal}). Therefore, we use our defined normal, which is the gradient of the $\sigma$ at point $x$, while calculating vertex colors. Fig.~\ref{lego} shows the advantage of our surface reconstruction method.

% 多视角融合的marching cube
% we extend NeRF to a generalized model and add surface constrain to model geometry varying geometry. To achieve the generalization of the model, we incorporate images feature from multi-views. Besides, as a result of $\sigma(\mathbf{x})$ discontinuity of opaque objects and high degrees of freedom of parameters in implicit fields, recovering details turn distorted or blurred. \\\\
% Bearing resemblance to Unisurf, $\sigma(\mathbf{x})$ is defined by the following function to solve this discontinuity problem:
% \begin{equation}
%     \sigma(x)=\left\{\begin{array}{l}>\tau, \text { if } x \text { inside the model} \\ =\tau, \text { if } x \text { on the surface} \\ <\tau, \text { if } x \text { outside the model } \end{array}\right.
% \end{equation}
% where $\tau$ is a hyperparameter used to represent the threshold of the density corresponding to the query point coordinate x inside and outside the object.
% Therefore, for each ray $\mathbf{r}$, we can determine if this ray intersects the target object by verifying $\sigma(\mathbf{r}(t))=\tau$. Practically, points at far and near distances contribute particularly differently to the color rendering. To this end, we modify the sampling strategy in NeRF: 1) sampling query points are only within the neighborhood $U\left(\mathbf{x}_{S}, \delta\right)$, where $\mathbf{x}_{S}$ denotes the intersection of ray and surface. 2) during training, the size of the neighborhood $\delta$ is decreasing. \\\\

\subsection{Separation of 3D Garment and Human Body}
Though we obtain the high-fidelity dressed person model $\mathcal{M}$ through the above surface reconstruction module, it is not enough for the limitation of the inseparable reconstruction of garments and the human body. For practical applications, such as virtual try-on and motion interaction, we propose a separating module to reconstruct garments and the human body individually.
% 这里下面的小标题不顶格，不确定需不需要将其顶格
\subsubsection{Parametric Body Reconstruction}
% 2D-3D mapping and confidence weighted
Existing separated reconstruction methods for dressed people \cite{mgn,corona2021smplicit,xiang2021modeling} estimate human body models by pose estimation from 2D images, which has a high degree of freedom and brings alignment issues to subsequent garments reconstruction. To avoid these problems, we propose the innovative USR pipeline, which first reconstructs the full 3D model, and then derives separated garments and the human body from the 3D model. In this way, the human body matches the full body model and does not require separate alignment with the garments. Specifically, the SMPL model of the human body is denoted by $\mathbb{B}(\beta,\theta;\Phi): \mathbb{R}^{10}\times\mathbb{R}^{3+3 K} \mapsto \mathbb{R}^{3N}$, where $\beta$ and $\theta$ respectively controls the shape and pose of the body mesh. 
% In addition, $\Phi$ contains 5 learned intrinsic parameters: $\Phi=(\overline{\mathbf{T}}, \mathcal{W}, \mathcal{S}, \mathcal{J}, \mathcal{P})$.
We assume that clothing is close to the body. Furthermore, we utilize the standard clothed human $\mathcal{M}$ as the ground truth to optimize the adjustable $\beta$ and $\theta$. Finally, the body reconstruction comes to solve the following optimization problem:
\begin{align}
\label{min}
\left(\beta^{*}, \theta^{*}\right)=\underset{\substack{\beta \in R^{10} \\ \theta \in R^{3+3 K}}}{\arg \min } \mathcal{L}_{C D}(U(B(\beta, \theta ; \Phi)), V(\mathcal{M})),
\end{align}
where $\mathcal{L}_{CD}$ is the Chamfer Distance between two point sets $U$ and $V$, which are sampled on SMPL body and the full dressed person mesh, responsively. For each point, Chamfer Distance 
finds the nearest point in another point set, and sums the
squared distances up normally. We remove the square of the distance calculation in the implementation to lift the stability of the algorithm. Concretely, for two-point sets $U$ and $V$, we use the $\mathcal{L}_{CD}$ as follows:
\begin{align}
\label{losscd}
\mathcal{L}_{CD}=\sum_{x\in U}\underset{y\in V}{\mathrm{min}}\left \| x-y \right \|_{2}+\sum_{y\in V}\underset{x\in U}{\mathrm{min}}\left \| y-x \right \|_{2}.
\end{align}
% where ${x\in U}$ and ${y\in V}$ are respectively the point
% sets down-sampled from vertices of generated mesh M and the ground truth points set. For each point, chamfer distance finds the nearest point in another point set, and sums the squared distances up.
Considering the optimization problem in Eq.(\ref{min}) is non-convex, we use SPIN \cite{spin} to estimate the initial values of beta and theta. Besides, in order to eliminate the modeling penetration phenomenon, a penetration loss term $\mathcal{L}_{\text {p}}$ is introduced to penalize vertexes outside $M(\beta, \theta ; \Phi)$:
\begin{align}
\label{pierce}
    \mathcal{L}_{\text {p}}=\sum_{v \in M(\beta,\theta; \Phi)} \sigma_{M}(v) d(v, \mathcal{M}),
\end{align}
\begin{align}
\label{disatnce}
    d(v, \mathcal{M}) \approx \min _{u \in \mathbf{V}_{M}}\|v-u\|_{2},
\end{align}
where the indicator function $\sigma_{\mathcal{M}}(v)$  takes the value 1 when $v$ is outside $\mathcal{M}$, and takes the value 0 otherwise. Consequently, the Iterative optimization algorithm in this section closes the performance gap between our model to the actual situation. Thus, the whole loss function of the regression of SMPL is:
\begin{align}
\label{loss}
\mathcal{L}=\mathcal{L}_{CD}(U,V)+\beta\mathcal{L}_{\text {p}}.
\end{align}
\subsubsection{Garments Reconstruction by SCGS}
%先说现有可分离方法基于偏移实现，天然地引入了拓扑约束，例如只能实现裤子，难以实现裙子，无法应对各种衣服的拓扑复杂性。因此我们提出基于分割的衣物重建方法，使得衣服重建主要依赖于具体图片信息，可以自由地实现各种拓扑的衣服。
% 这个模块需要加入那张投影的图片吗？
Most garments reconstruction methods deform the garments templates to derive the 3D garments \cite{corona2021smplicit}, which requires knowing the type of clothes in advance and training multiple reconstruction models independently with 3D supervision. On account of the fixed topology, template-based approaches limit the flexibility to deal with various complex garments. For instance, reconstructing a skirt on a pants template leads to an incorrect model for they share different topologies. Moreover, high cost for 3D supervision signal acquisition goes against the training. To solve these problems, we propose a segmentation-based garment reconstruction method, with a Semantic and Confidence Guided Separation strategy (SCGS) to detect, segment, and reconstruct the clothes layer, without any 3D label or model. SCGS makes full use of the pixel-level and semantic-level characteristics of 2D images and fuses multi-view corresponding information with confidence, to complete the extraction of the 3D garments via the geometric and semantic consistency between 2D and 3D.

First, we utilize a pre-trained 2D segmentation network Cloth-seg \cite{clothseg} to perform 2D segmentation on the input multi-view images, labeling pixels with Upper, Lower, Fully, or Non-Clothing categories. The semantic segmentation result of each pixel is denoted as a 4D vector $\mathbf{c_{lk}}$, representing the probability that the pixel belongs to the four categories. Then, we perform semantic segmentation on all vertices of the 3D mesh using SCGS, shown in Fig.~\ref{fig1}.

Specifically, given a mesh model $\mathcal{M}$ and a 2D semantic segmentation map $s_{i}$ at the $\mathrm{i}^{\text {th}}$ perspective, we project vertex $v_{j}$ of $\mathcal{M}$ onto $s_{i}$ to get the projection point $p_{j}^{i}$.
Then, for every projection point, we apply the bilinear interpolation method on its nearest four pixels confidence to determine the semantic segmentation confidence of the projected point, denoted by $\mathbf{c}_{p_{j}^{i}}$.
Finally, for each vertex $v_{j}$ on $\mathcal{M}$, we average the $\mathbf{c}_{p_{j}^{i}}$ obtained from all perspectives to estimate the 3D segmentation confidence $\mathbf{c}_{v_{j}}$:
\begin{align}
\label{3dpconfidence}
    \mathbf{c}_{v_{j}}=\frac{\sum_{i=1}^{N} \delta_{j}^{i} \mathbf{c}_{p_{j}^{i}}}{\sum_{i=1}^{N} \delta_{j}^{i}}.
\end{align}
Note that $\delta_{j}^{i}$ is an indicator function that takes 1 if vertex $v_{j}$ can be observed by view $\mathrm{i}^{\text {th}}$; otherwise takes 0.
% To ameliorate the segmentation strategy, we additionally filter out unreliable perspectives, whose number of non-clothing pixels is below a certain threshold. As evidenced by the extensive  experiments, our method achieves high fidelity and high-quality details.
\subsection{Geometry Fine-tune Module}
\begin{figure}[htbp]
\centering
\includegraphics[width=7.5cm]{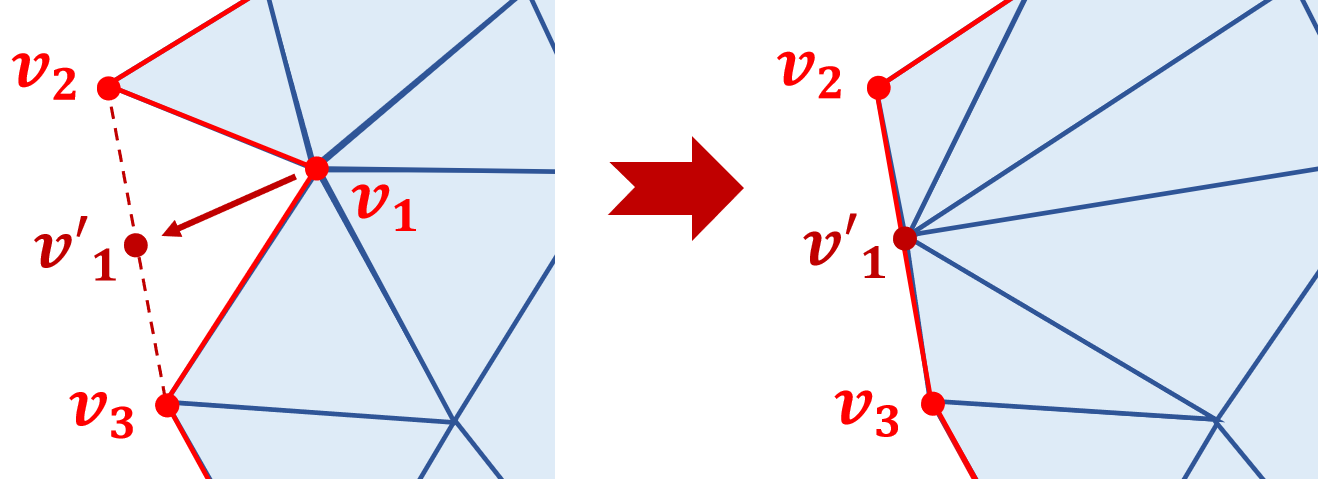}
%\setlength{\belowcaptionskip}{-0.5cm}
%\vspace{-0.25cm}
\caption{Illustration of the edge smoothing algorithm.}
\label{topo}
\end{figure}

To refine the reconstructed garments meshes, we introduce a geometry fine-tune module to further refine edge details, illustrated in the Fig.~\ref{topo}. The edge smoothing algorithm mainly includes two steps: edge detection and edge smoothing. First, by recording the number of faces that each edge is involved in, we find a set of garment edges $\mathbf{E}_{\text {border }}$ and a set of border points $\mathbf{V}_{\text {border }}$, which are endpoints of border edges. Second, we randomly select three consecutive points $v_{1}$, $v_{2}$, $v_{3}$ according to the connected relationship, and directly move $v_{1}$ to the midpoint of $v_{2}$ and $v_{3}$:
\begin{align}
\label{finetune}
   v_{1}^{\prime}=\frac{1}{2}\left(v_{2}+v_{3}\right).
\end{align}
Note that rounds of edge smoothing operation are not in direct proportion to the quality. Without special instruction, we perform 5 rounds of edge smoothing operation by default.

\subsection{Parametric Dressing Model SMPL-D}
%半页
Focusing on the application of our model, we propose the SMPL-D model for the convenient interaction of body and clothing. 
The SMPL-D model is a SMPL-based dressed characterization model, indicating the correspondence between clothed body model and SMPL model.
For the arbitrary clothed body model in any static scene, we first reconstruct separated human and garments models. Then, for every vertex $\mathbf{v}$ in garments models, we find its smallest offset to the vertex $\mathbf{u}_{\mathbf{v}}$ in SMPL model. Consequently, this set of correspondences and offsets is recorded as $\mathcal{O}(\mathbf{v}_{i})$:
\begin{align}
    \mathcal{O}\left(\mathbf{v}_{i}\right)=\left(\mathbf{u}_{\mathbf{v}_{i}}, \mathbf{v_{i}}-\mathbf{u}_{\mathbf{v}}\right), i=1,2, \cdots, m,
\end{align}
where $m$ represents the number of vertexes in garments models. According to the set $\mathcal{O}$, it is easy to register the reconstructed 3D garments on the other body model. Additionally, SMPL-D guarantees that garments in the new clothed model maintain the same topology and similar appearance. However, the shape of garments changes due to different body shapes.

\section{Experiments}
\subsection{Settings}
\subsubsection{Dataset} 3D dressed people datasets with ground truth measurements are rare. Renderpeople \cite{Renderpeople} is the most popular dataset used in dressed people reconstruction methods \cite{saito2019pifu,saito2020pifuhd}, but it is expensive for the complex production process and huge commercial value, bringing difficulties to the experiments. Therefore, we propose an Open-source Diverse Dressed People dataset (ODP), which contains 25 dressed people models and will be continuously expanded. In our experiment, we randomly split the models into 15, 5, and 5 for training, validation, and test, respectively. There is no intersection between training and test sets, ensuring the validity of generalization tests. For every 3D model, we render images of $512\times512$ resolutions from 140 random views. In the training stage, in order to realize data augmentation, we randomly choose 20 images from the 140 images as input for each epoch. In the test stage, we still infer a 3D model from images of 20 random views for every test person. We also keep the same settings for the comparison methods.

\begin{figure*}[ht]
\centering
\includegraphics[width=18cm,height=8cm]{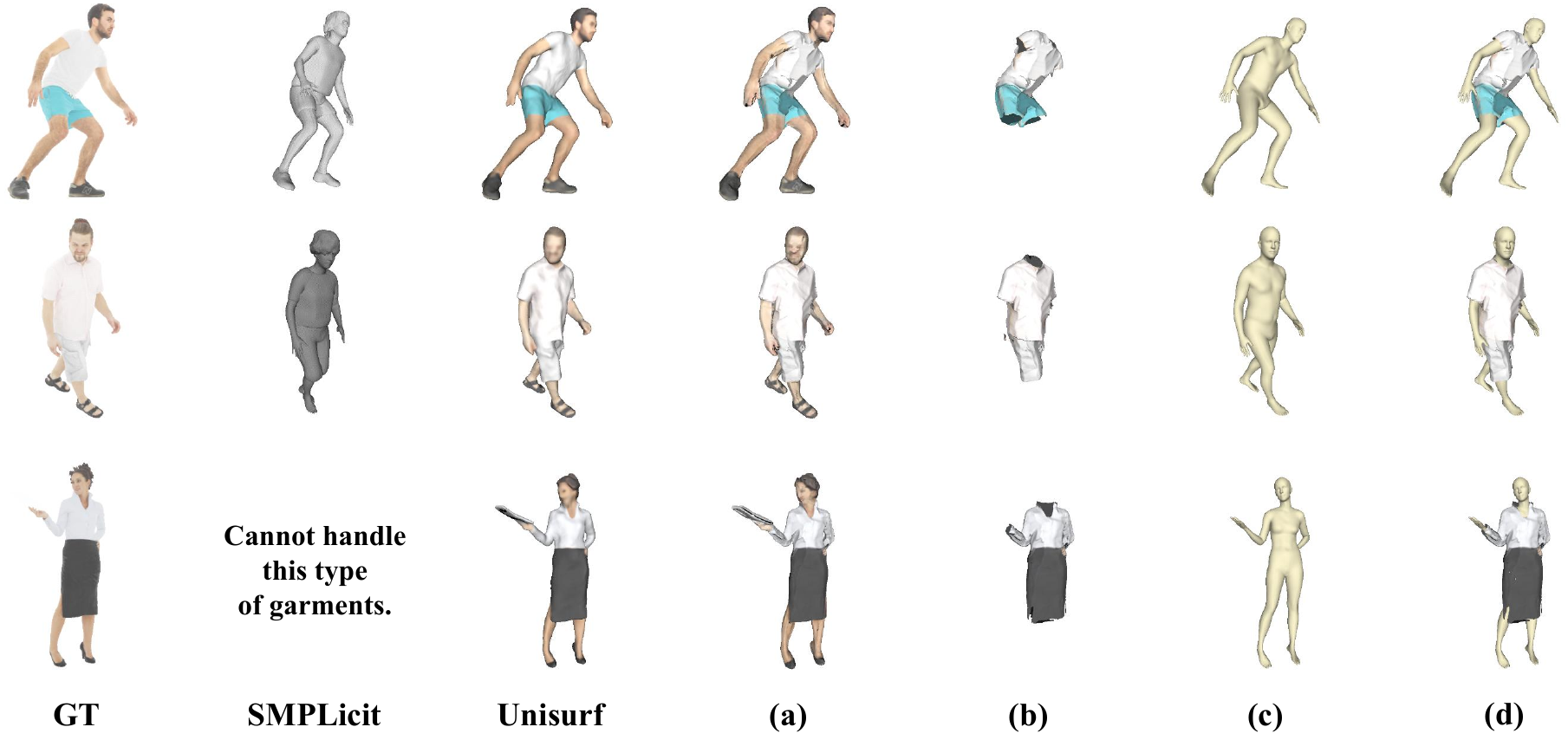}
\setlength{\belowcaptionskip}{-0.5cm}
\vspace{-0.5cm}
\caption{Qualitative comparison between ours and the state of the
arts.}
\label{result1}
\end{figure*}

\begin{table}[t]
\centering
\resizebox{.48\textwidth}{!}{
\setlength\tabcolsep{1.2pt}{
\renewcommand\arraystretch{1.8}
\begin{tabular}{m{2.5cm}<{\centering}|m{2cm}<{\centering}m{2cm}<{\centering}|m{3cm}<{\centering}}
%\toprule
\hline\hline
Metric & \multicolumn{2}{c|}{Appearance} & Geometry  \\
%\midrule
\hline
Method & PSNR $\uparrow$ & SSIM $\uparrow$ & Chamfer Distance $\downarrow$ \\
\hline
SMPLicit & {$-$} & $-$ & {\makecell[c]{0.0197\\(without fail cases)}} \\
{\makecell[c]{Unisurf\\(per-scene opt)}} & 21.878 & \textbf{0.859} & 0.00039 \\
Ours & \textbf{24.256} & 0.826 & \textbf{0.00028}  \\
%\bottomrule
\hline\hline
\end{tabular}
}
}
\vspace{-0.2cm}
\caption{Quantitative comparison of the appearance and the geometry. The evaluation metrics are PSNR (higher is better), SSIM (higher is better) and Chamfer Distance (lower is better).}
\label{compareresults}
\vspace{-0.5cm}
\end{table}

\subsubsection{Evaluation Metrics} We validate the efficacy of our model in terms of both appearance realism and geometric accuracy. For the quantitative evaluation of the appearance, we render the obtained 3D colored mesh to images of 14 random views and calculate the image quality using PSNR and SSIM metrics. For geometric accuracy, we use the Chamfer distance \cite{3drecon} to evaluate the differences between the predicted 3D model and the ground truth, which sums the squared distances between nearest neighbor correspondences of two-point clouds sampled on the meshes. 

\subsubsection{Implementation Details} For the image encoder, we adopt the U-ResNet34. We organize the batch using the sampling points on the rays and utilize the Pointnet-like MLPs to decode the features of points to the color and density. We train the model using the training dataset and test it on unseen people. We use Adam \cite{kingma2014adam} optimizer with an initial learning rate of $5\times {10}^{-4}$, which decays exponentially along with the optimization. We train our network using one RTX 3090 Ti GPU for 10000 epochs, spending about 20 hours. The values of hyper-parameters mentioned above are set as $\lambda=0.5, \beta=1$.

In this section, we show the results of our methods qualitatively and quantitatively, and compare them with the state-of-the-art methods. Since we are the first unsupervised method to separately reconstruct dressed people from images, we choose SMPLicit \cite{corona2021smplicit} and Unisurf \cite{oechsle2021unisurf} for the comparison, both of which are the state-of-the-art methods, respectively being selected from supervised separated reconstruction methods and unsupervised inseparable non-generalizable reconstruction methods. To be fair, we optimize SMPLicit with multiple images, and optimize Unisurf on all test data one scene by one within the same training time. We also show a real-world application in the form of trying on garments recreated from our models for people of different sizes and genders.

\subsection{Results}

\subsubsection{Qualitative results}
As the Fig.~\ref{result1} shows, Our USR model reconstructs the full people models, 3D garments and human bodies with high fidelity in both appearance-level and geometry-level. The reconstructed results of SMPLicit are far from the ground truth and lack of geometry details, such as folds and fine structures like vest straps. Besides, SMPLicit cannot deal with color and texture.   By contrast, Unisurf achieves acceptable results, but it needs to be trained for one day a person, having no capability of generalization and separation of garments and the body. More examples are shown in the supplementary material.

\subsubsection{Quantitative results} The quantitative results are shown in Tab.~\ref{compareresults}. As we can see, our method achieves the lowest Chamfer Distance and PSNR comparing with other methods. Also, the SSIM of out method is pretty close to Unisurf, which is a per-scene optimization method and ought to perform better than generalizable methods. These results validate the effectiveness of our proposed GSNeRF.

\begin{figure*}[h]
\centering
\includegraphics[width=17cm]{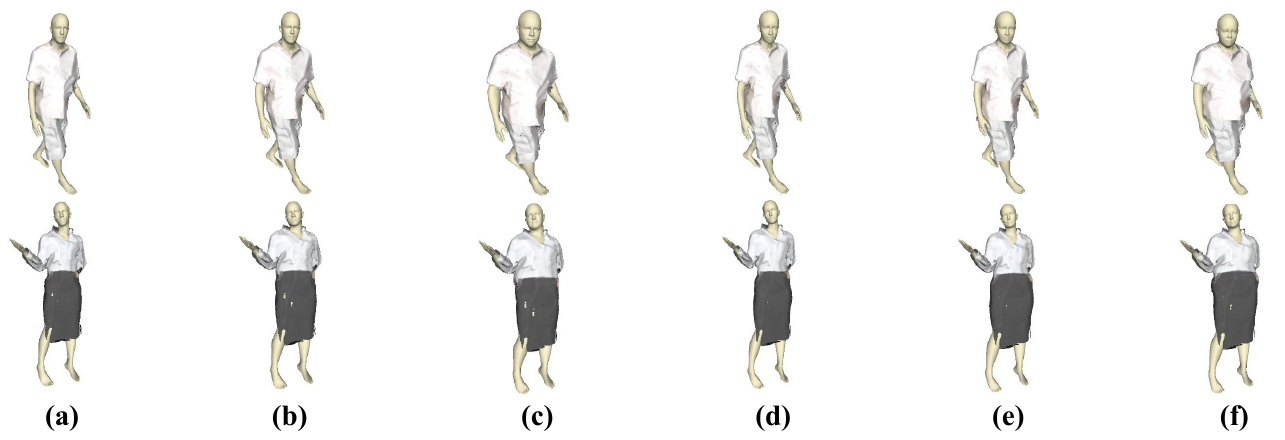}
\setlength{\belowcaptionskip}{-0.5cm}
\vspace{-0.25cm}
\caption{Virtual try-on results for people with different body types and different genders.}
\label{tryonresults}
\end{figure*}

% \begin{figure}[htdp]
% \centering
% \includegraphics[width=6cm]{Fig/n.pdf}
% %\setlength{\belowcaptionskip}{-0.5cm}
% \caption{Try-on results with increasing $\lambda$.}
% \vspace{-0.8cm}
% \label{normalweightfig}
% \end{figure}

\subsubsection{Application on the Try-on}
As concisely mentioned above, we are capable of building SMPL-D based on reconstructed SMPL and garments models. To verify this functionality, we design the following experiments: for an arbitrary character model in our dataset, with reconstructed SMPL parameters, we reconstruct a SMPL model with the same pose as the original one but with a different shape; then we apply the prepared SMPL-D model to "dress" the body. 
Intuitively, as shown in Fig.~\ref{tryonresults}, we realize virtual try-on between different body types and even different genders. Such results prove the great role of separation reconstruction for human reconstruction, and separation reconstruction will have broader application scenarios than integral reconstruction in the future.

\subsection{Ablation Studies}
In this section, we compile a set of ablation experiments to verify the importance of two components in our method, including the normal loss in the generalized surface-aware neural radiance field (GSNeRF), the use of gradients in surface reconstruction, and the fine-tune module. 

\subsubsection{The analysis of normal loss} For the analysis of the normal loss, we set different weight for it and reports the surface reconstruction results as Tab.~\ref{ablation}. We show the qualitative results in the supplementary materials. When there is no normal constraint in the radiance field, the reconstructed results are rough. As the weight of normal loss increases, the results become smoother. Because we utilize the gradient of the occupancy as the normal of each 3D point during the explicit surface generation, the results of (b) and (c) have some black faces, due to the reverse direction of the normal direction at some points. This phenomenon is eliminated when the normal constraint is increased, which corrects the wrong normal. When the weight of the normal constraint continues to increase, the surface is too smooth and loses detail. 
% \begin{figure}[htbp]
% \centering
% \includegraphics[width=8.5cm]{Fig/topo.png}
% %\setlength{\belowcaptionskip}{-0.5cm}
% %\vspace{-0.25cm}
% \caption{Illustration of the edge smoothing algorithm.}
% \label{topo}
% \end{figure}
We also show the changes of quantitative metrics while the weight of the normal loss is increasing. The quality of the rendered image is the worst when there is no normal constraint, and the best when the weight of the constraint is 0.1. Taking into account the indicator results and visual effects, we choose the weight of normal loss equal to 0.1 in our experiments. 

\begin{figure}[htbp]
\centering
\vspace{-0.25cm}
\includegraphics[width=4.5cm]{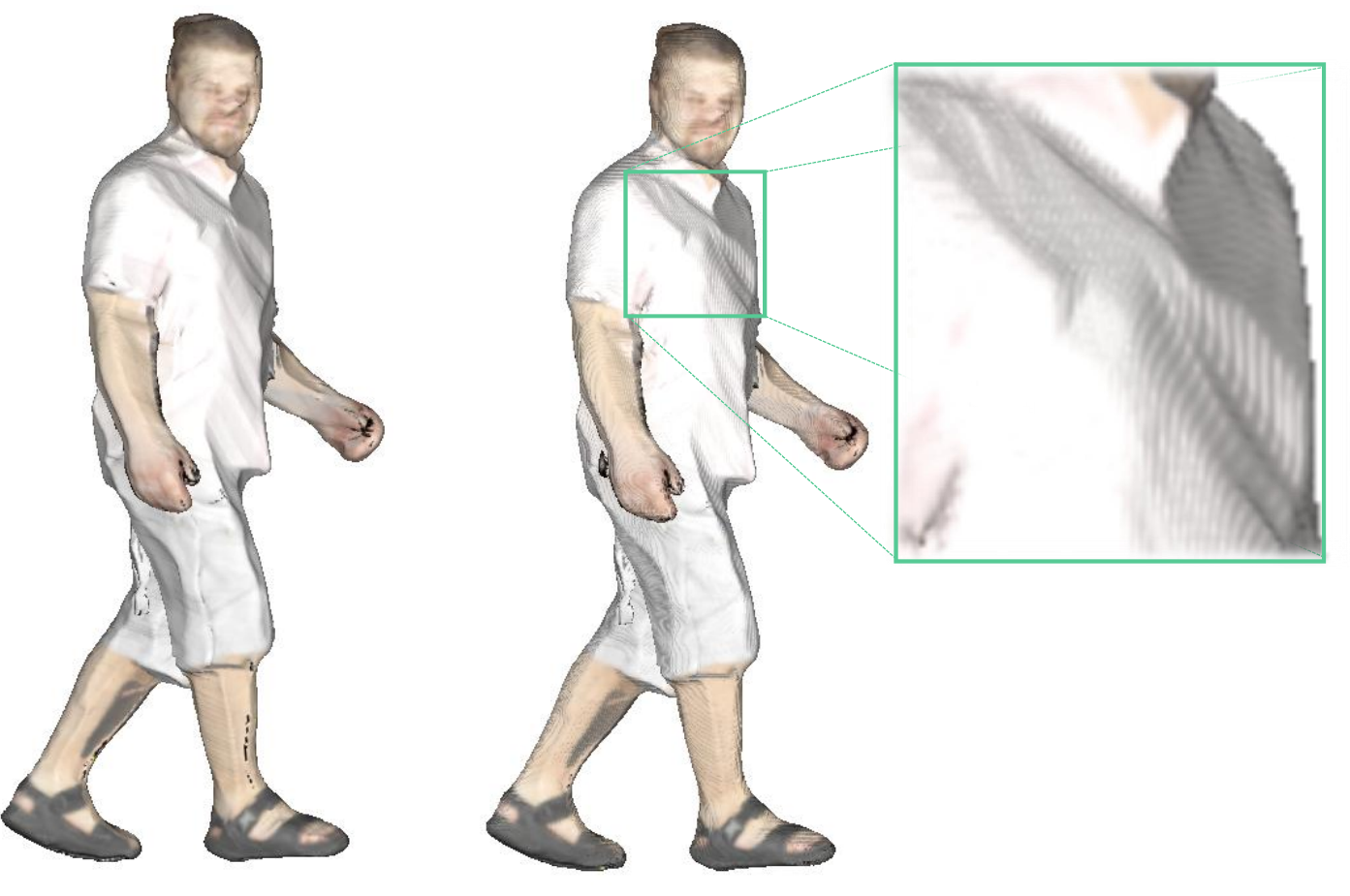}
\vspace{-0.25cm}
\caption{The comparison of the reconstruction results with and without the normal calculated by the gradient of occupancy.}
\label{lego}
\vspace{-0.2cm}
\end{figure}
\vspace{-0.2cm}

\subsubsection{The analysis of the surface normal} 
Marching Cube \cite{marchingcube} algorithm is a classic iso-surface extraction algorithm. In our experiments, we found that the reconstructed 3D meshes have obvious graininess if we calculate the normal of points according to the neighbor points. Although increasing the resolution of the cube and paying a larger computational and time penalty will result in slightly smoother results, it still does not eliminate this phenomenon. Instead, when we use the gradient direction of the occupancy at the point as the normal of this point, the derived surface by marching cube is smoother under the same cube resolution. The reason is that the normal calculated by the gradient has experienced the optimization of the normal loss in Eq.(\ref{normal}). 

% \subsubsection{The analysis of the Fine-tune Module}
% We show the change of local details before and after the edge smoothing operation in the Fig.~\ref{}. The Fine-tune Module shows great advantages in removing burrs and smoothing the edges of garments. Our proposed Fine-tune Module is extensible and easily migrated to all application scenarios that require mesh post-processing.

\begin{table}[t]
\centering
\resizebox{.47\textwidth}{!}{
\setlength\tabcolsep{1pt}{
\renewcommand\arraystretch{1.8}
\begin{tabular}{m{2.8cm}<{\centering}|m{1.2cm}<{\centering}|m{1.2cm}<{\centering}|m{1.2cm}<{\centering}|m{1.2cm}<{\centering}|m{1.2cm}<{\centering}}
%\toprule
\hline\hline
weight & 0 & 0.005 & 0.05 & 0.1 & 0.5  \\
%\midrule
\hline
Chamfer Distance$\downarrow$ & 00080  & 0.00043 & 0.00030 & 0.00028 & 0.00052\\
\hline\hline
%\bottomrule
\end{tabular}
}
}
\vspace{-0.15cm}
\caption{Ablation study of the weight of the normal loss.}
\label{ablation}
\vspace{-0.5cm}
\end{table}

\section{Conclusion}
In this paper, we present an unsupervised separated reconstruction framework (USR)  with a generalized surface-aware neural radiance field (GSNeRF) and a strategy and a Semantic and Confidence Guided Separation strategy (SCGS) to separately reconstruct the 3D garments, and the human body from only multi-view images. Extensive experiments have demonstrated that USR is capable of generating 3D models with high-fidelity geometry and appearance for unseen people. In future work, we will explore reconstructing methods for less inputs. 
\bibliographystyle{aaai23} 
\bibliography{aaai23}

\end{document}